\documentclass[conference]{IEEEtran}
\def\IEEEtitletopspace{18pt}
\IEEEoverridecommandlockouts
\usepackage{ifpdf}
\ifpdf
  \usepackage{hyperref}
  \usepackage{graphicx}
\else
  \usepackage[dvipdfmx]{hyperref}
  \usepackage[dvipdfmx]{graphicx}
\fi
\usepackage{amsfonts}           
\usepackage{cite}               
\usepackage{amsmath,amssymb,bm} 
\usepackage{mathtools}          
\usepackage{siunitx}            
\usepackage{nicefrac}           
\usepackage{booktabs}           
\usepackage{subcaption}         
\def\BibTeX{{\rm B\kern-.05em{\sc i\kern-.025em b}\kern-.08em
    T\kern-.1667em\lower.7ex\hbox{E}\kern-.125emX}}
\begin{document}

\title{Transformation Categorization Based on Group Decomposition Theory Using Parameter Division
}

\author{
Takayuki Komatsu$^{1}$, Yoshiyuki Ohmura$^{1}$, and Yasuo Kuniyoshi$^{1,2}$
\thanks{
$^{1}$Graduate School of Information Science and Technology, The University of Tokyo, Tokyo, Japan.
$^{2}$Next Generation Artificial Intelligence Research Center (AI Center), The University of Tokyo, Tokyo, Japan.
{\tt\small \{komatsu, ohmura, kuniyoshi\}@isi.imi.i.u-tokyo.ac.jp}
This study was supported by JSPS KAKENHI (25H00448), Japan.
The funding sources had no role in the decision to publish or prepare the manuscript.
}
}

\maketitle

\begin{abstract}
Representation learning aims to acquire meaningful representations from sensory inputs without supervision.
This approach can serve as a computational model of human development.
Although many neural network models have been proposed that empirically learn meaningful representations, no principled formulation of what constitutes a good representation exists.
We have been studying representation learning theory under algebraic structural constraints, and within this framework we develop methods that categorize changes between pairs of sensory inputs.
Conventional representation learning aims to acquire mutually independent representations.
However, representations exist that are not mutually independent and these methods do not apply to them.
To address this problem, we recently introduced Galois algebra theory, which decomposes a group using its normal subgroups.
This method learns to decompose individual transformations into the product of two transformations while constraining either transformation to belong to a normal subgroup.
This method is more general than conventional approaches because it is applicable to both independent and non-independent categorization.
However, it relies on several auxiliary assumptions that are unrelated to group decomposition theory.
Such assumptions unduly narrow the range of scenarios to which the method can be applied.
Furthermore, no ablation study has been conducted to determine whether appropriate categorization is achieved by the theory-based or auxiliary constraints.
In this study, we propose an alternative formulation that is not a decomposition into the product of two transformations.
In this method, we parametrize the single transformation by multiple parameters.
Then, we impose constraints based on group decomposition theory on one of the parameters.
We obtain the normal subgroup as the set of transformations that result when that parameter is fixed to the identity element.
As a result, the proposed method avoids the auxiliary assumptions in the previous study and applies to a broader range of scenarios.
Through an ablation study, we showed that appropriate categorization is achieved when the theory-based constraints are imposed.
\end{abstract}


\section{Introduction}\label{sec:introduction}

Humans extract essential information such as object size, position, and orientation from sensory input.
These abilities develop during infancy.
Representation learning is a branch of artificial intelligence that seeks to learn meaningful representations of sensory input without supervision.
This approach can serve as a computational model of human development \cite{Takada2021, Nishitsunoi2024, Komatsu2025}.

In early work, researchers suggested that a good representation should consist of disentangled components \cite{Bengio2013}.
Many existing methods emphasize statistical independence among scalar dimensions \cite{Higgins2017, Chen2016, Yang2023}.
Although they learn quantities such as object size and position along a single axis, they often over-split one factor into multiple dimensions or collapse multiple factors into one.
In later work, Higgins et al.\ \cite{Higgins2018} offered a more general, algebra-based definition of disentanglement in terms of symmetric transformations, but did not provide a specific learning procedure, leaving the link between theory and practice open.

To address this gap, we proposed learning to categorize changes between sensory inputs under algebraic structural constraints.
Then, we considered algebraic independence \cite{Ohmura2025, Nishitsunoi2024}, a mathematical generalization of independence \cite{Simpson2018} whose key requirement is commutativity: the order of applying two transformations does not matter, that is, $a \circ b=b \circ a$.
Our methods successfully categorize transformations into independent vector dimensions, such as color and shape.
However, not all natural representations for humans satisfy independence.
For example, translation and rotation do not commute because the order of these transformations affects the center of rotation.
To also address such cases, a categorization criterion is required that differs from commutativity yet remains applicable when commutativity holds.

To address this problem, we recently proposed a method \cite{Nishitsunoi2025} based on group decomposition in Galois theory \cite{Singh1999}, which generalizes commutativity.
A group is a set structured by a binary operation.
Diverse symmetries are known to admit description in group-theoretic terms.
In group decomposition theory, a normal subgroup $N$ of $G$ and a map $f$ from $G$ to another group $H$ play central roles.
A \emph{homomorphism} $f \colon G \to H$ is a map that preserves the structure of the group operation on $G$; a normal subgroup is obtained through such a map.
That normal subgroup $N$ then decomposes $G$.
To implement this theory, the method of \cite{Nishitsunoi2025} builds on two of our previous studies: (i) decomposing a single transformation into a product of two factors \cite{Ohmura2025} and (ii) formulating object transformations in image sequences \cite{Takada2022}.
For a sequence of $T$ frames $\mathbf{x}_0, \ldots, \mathbf{x}_{T-1}$, the transformation from $\mathbf{x}_0$ to $\mathbf{x}_i$ is formulated as $\mathbf{x}_i = (v_i \circ g_i) \mathbf{x}_0$, where $g_i$ and $v_i$ denote the two factor transformations.
While assuming that each transformation $(v_i \circ g_i)$ lies in $G$, that method learns transformations $g_i$ and $v_i$ so that the map $f \colon (v_i \circ g_i) \mapsto v_i$ is a homomorphism.
As a result, object motion that includes rotation and translation is treated as a non-commutative case, and translation is successfully extracted as the normal subgroup.

However, that method relies on auxiliary assumptions that are not part of group decomposition theory.
The first is the assumption of uniform linear motion: the transformation $(v_i \circ g_i)$ from $\mathbf{x}_0$ to $\mathbf{x}_i$ is expressed using the transformation $(v_1 \circ g_1)$ from $\mathbf{x}_0$ to $\mathbf{x}_1$, but in the form $(v_i \circ g_i) = v_1^i \circ g_1^i$ rather than $(v_i \circ g_i) = (v_1 \circ g_1)^i$.
The second is the assumption that the factor transformations $g_i$ and $v_i$ are isometric, which rules out more complex, non-isometric solutions such as scale and shear.
Furthermore, we did not conduct an ablation study in \cite{Nishitsunoi2025}; hence, we did not show that appropriate categorization is achieved by the theory-based constraints rather than the auxiliary assumptions alone.
Thus, the limitations of that work are the narrow applicability imposed by the auxiliary assumptions and the lack of experimental verification.

To overcome these limitations, we focus on the flexibility of the original mathematical formulation of group decomposition theory.
In our previous study, we considered the map $f \colon (v \circ g) \mapsto v$, where $v \in H$ and $(v \circ g) \in G$.
However, this implies that, in the previous study, we imposed two overly strong assumptions that are not theoretically required by the original formulation: (1) the binary operation on $H$ coincides with that on $G$ and (2) $H$ is a subgroup of $G$.
Therefore, when we introduce group decomposition theory into transformation categorization, there is room to reconsider the framework of decomposing each transformation into a product of two transformations.

In this study, we propose a framework that divides the parameter of a single transformation into multiple parts.
We denote the parameter by $\theta$ and multiple components $\phi_1,\ldots,\phi_n$, write the transformation as $g(\theta, \phi_1,\ldots,\phi_n)$, and assume the map $f \colon g(\theta, \phi_1,\ldots,\phi_n) \mapsto \theta$.
Then, we impose constraints so that $f$ is a homomorphism.
Using this constraint, we acquire the normal subgroup without the auxiliary assumptions we relied on in the previous study.
Thus, our method handles a broader class of cases than that in the previous study, where auxiliary assumptions had limited applicability to transformations such as scale transformation.

We validate the proposed method on pairs of images of object transformations that include rotation, translation, and scale.
Using an ablation study, we show that the proposed constraints based on group decomposition theory achieve appropriate transformation categorization.
Our model contributes to a model of how humans develop a recognition of various object transformations.

\section{Transformation Categorization}

In this section, we outline transformation categorization based on group decomposition theory using parameter division, focusing on the mathematical structure and omitting the implementation details.

\subsection{Definition of group}

A \emph{group} is a pair of a set $G$ and a binary operation $\circ$ on $G$ that satisfies the following four axioms: (1) closure, (2) associativity, (3) existence of an identity element, and (4) existence of an inverse element:
\begin{equation}
    g_1 \circ g_2 \in G, \quad \forall g_1, g_2 \in G,
\end{equation}
\begin{equation}
    (g_1 \circ g_2) \circ g_3 = g_1 \circ (g_2 \circ g_3), \quad \forall g_1, g_2, g_3 \in G,
\end{equation}
\begin{equation}
    \exists e \in G, \quad e \circ g = g \circ e = g, \quad \forall g \in G,
\end{equation}
\begin{equation}
    \forall g \in G, \; \exists g^{-1} \in G, \quad g \circ g^{-1} = g^{-1} \circ g = e.
\end{equation}

\subsection{Normal subgroups}

A subgroup $N$ of $G$ is a \emph{normal subgroup} if $gN = Ng$ for all $g \in G$, where $gN$ and $Ng$ are defined as follows:
\begin{align}
    gN &= \{g \circ n \mid n \in N\}, \quad \forall g \in G, \\
    Ng &= \{n \circ g \mid n \in N\}, \quad \forall g \in G.
\end{align}

If the operation is commutative, then $g \circ n = n \circ g$.
By contrast, normality is characterized by the condition $gN = Ng$ and $g \circ n \neq n \circ g$ is allowed.
Thus normal subgroups generalize commutativity.
The normal subgroup $N$ yields a well-defined partition of $G$, which Galois referred to as a proper decomposition \cite{Singh1999}.

Normal subgroups can be obtained from group homomorphisms.
A \emph{homomorphism} $f \colon G \to H$ is a map between groups $G$ and $H$ that satisfies
\begin{equation}
    f(g_1 \circ g_2) = f(g_1) \cdot f(g_2), \quad \forall g_1, g_2 \in G,
    \label{eq:group_homomorphism}
\end{equation}
where $\cdot$ is the binary operation on $H$.
The \emph{kernel} of $f$, denoted by $\operatorname{Ker}(f)$, is a normal subgroup of $G$:
\begin{equation}
    \operatorname{Ker}(f) := \{g \in G \mid f(g) = e_H\},
    \label{eq:group_kernel}
\end{equation}
where $e_H$ is the identity element of $H$.

From the viewpoint of representation learning, this framework is characteristic in that it can be interpreted as hierarchical structure learning, not merely as a decomposition.
In most previous studies, researchers focused on representation decomposition based on relationships among lower-level categories, and did not explicitly model the relationship between higher-level categories and their lower-level subcategories.
By contrast, learning a normal subgroup (i.e., learning $N \subset G$) explicitly addresses the relationship between the higher-level category $G$ and the lower-level category $N$ through the homomorphism $f$.

\subsection{Formulation of transformation categorization}\label{sec:transformation_categorization}

We consider a transformation $g$ from one sensory input $\mathbf{x}$ to another sensory input $\mathbf{y}$ and assume that this transformation belongs to a group $G$.
Our goal, as a representation-learning objective, is to obtain a normal subgroup $N$ of $G$ and thereby categorize transformations in $G$ via the induced partition by $N$.

To define the homomorphism $f$ of $G$, we parameterize the transformation $g$ by multiple components $\theta, \phi_1,\ldots,\phi_n$.
Then the transformation is written as
\begin{equation}
    \mathbf{y} = g(\theta, \phi_1,\ldots,\phi_n) \mathbf{x}.
    \label{eq:transformation_param}
\end{equation}
We assume a map $f \colon G \to H$ from the group $G$ to another group $H$ defined as
\begin{equation}
    f(g(\theta, \phi_1,\ldots,\phi_n)) = \theta.
    \label{eq:define_f}
\end{equation}

When $f$ is a homomorphism, from Eq.~\ref{eq:group_homomorphism} and Eq.~\ref{eq:define_f}, we obtain the constraint on the parameter $\theta$.
We define $g_1=g(\theta_1, ...), g_2=g(\theta_2, ...), g_{1,2}=g_1 \circ g_2=g(\theta_{1,2}, ...)$.
Then, we obtain
\begin{equation}
    \theta_{1,2} = \theta_1 \cdot \theta_2.
    \label{eq:homomorphism_constraint}
\end{equation}
Therefore, we constrain the parameter $\theta$ to satisfy this law.

\begin{figure}[t]\centering\includegraphics[trim=110 48 180 80, clip, width=\linewidth]{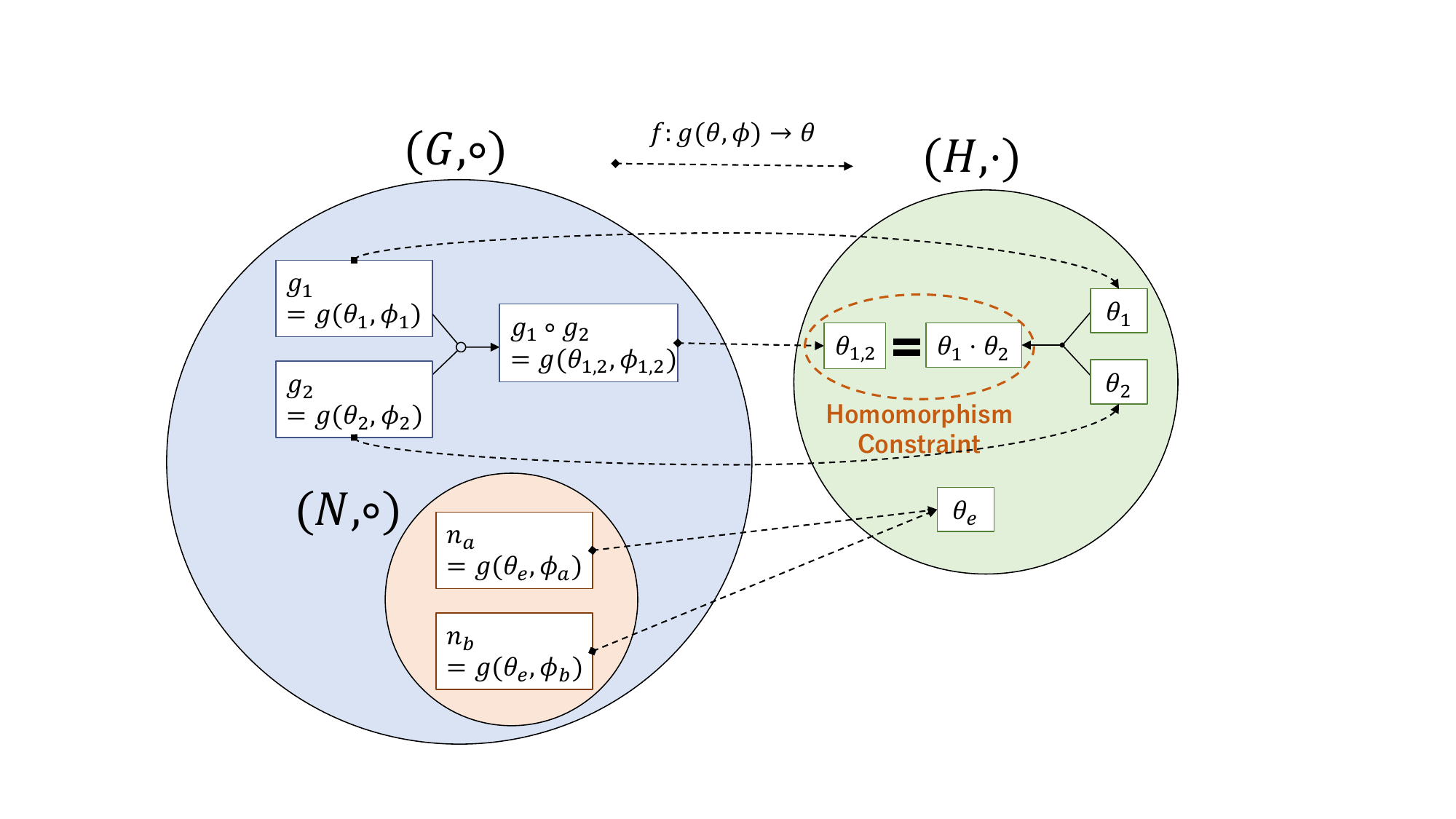}\caption{
Schematic of the proposed formulation in the case of two parameters.
By the closure property of a group $G$, when this group contains $g_1=g(\theta_1, \phi_1)$ and $g_2=g(\theta_2, \phi_2)$, there exists $g_{1,2}=g(\theta_{1,2}, \phi_{1,2})$ which satisfies $g_{1,2}=g_1 \circ g_2$.
$g_1$, $g_2$, and $g_{1,2}$ are mapped to $\theta_1$, $\theta_2$, and $\theta_{1,2}$, respectively, by the mapping $f \colon g(\theta, \phi) \mapsto \theta$.
If $f$ is a homomorphism, $\theta_{1,2}=\theta_1 \cdot \theta_2$ should hold.
Using this homomorphism $f$, we can obtain the normal subgroup $N$ of $G$ as the set of $g$ under the condition that $\theta$ is set to the identity element $\theta_e$.
}\label{fig:1}\end{figure}

Finally, we acquire the normal subgroup $N$ as the kernel of $f$:
\begin{equation}
    \operatorname{Ker}(f) = \{g(\theta_e, \phi_1,\ldots,\phi_n) \mid \phi_1,\ldots,\phi_n\},
\end{equation}
where $\theta_e$ is the identity element of $H$.
We show the overview of the proposed formulation in the case of two parameters in Fig.~\ref{fig:1}.

In our previous study \cite{Nishitsunoi2025}, we made two assumptions: (1) the binary operation on $H$ coincides with that on $G$, and (2) $H$ is a subgroup of $G$.
By contrast, we do not impose these assumptions on the proposed method.
The design of $\theta$ and the corresponding binary operation $\cdot$ do not require the above two constraints, which provides greater flexibility.

\section{Learning Model}\label{sec:learning_model}
In this section, we describe how to implement the formulation in the previous section using neural networks (NNs).

\subsection{Object transformation}

As in our previous study \cite{Nishitsunoi2025}, we focus on geometric transformations of an object.
We describe the key modeling choices; for full details, refer to \cite{Nishitsunoi2025}.

We consider a transformation $g$ between two images $\mathbf{x}$ and $\mathbf{y} \in \mathbb{R}^{H \times W \times 3}$.
We formulate geometric transformations of the object as coordinate shifts of pixel values.
Let $\mathbf{p} = (x, y)^\top \in \mathbb{R}^2$ be the input pixel position, $\Delta \mathbf{p} = (\Delta x, \Delta y)^\top$ be its displacement, and $\mathbf{p}' = \mathbf{p} + \Delta \mathbf{p}$ be the output pixel position.
We obtain the transformed image $g\mathbf{x}$ by resampling the input image $\mathbf{x}$ based on the input and output pixel positions $\mathbf{p}$ and $\mathbf{p}'$, respectively.
We implement the actual pixel-wise transformation process using spatial transformer networks \cite{Jaderberg2015}, as in \cite{Nishitsunoi2025}.

\subsection{Parameterization of the transformation $g$}

We parameterize $g$ using $(\theta, \phi_1, \ldots, \phi_n)$, as in Section~\ref{sec:transformation_categorization}, where $\theta \in \mathbb{R}^{d_\theta}$ and $\phi_i \in \mathbb{R}^{d_{{\phi},i}}$ are vectors.
In this section, for simplicity, we focus on the case $n=1$ and denote $\phi_1$ by $\phi \in \mathbb{R}^{d_\phi}$.
The transformation $g(\theta, \phi)$ consists of two NN modules.
We can view $g$ as consisting of two modules: (1) a displacement-producing module that takes an input and output pixel position $\mathbf{p}$ and $\mathbf{p}'$, and (2) an pixel-wise transformation module that actually transforms the image based on the input and output pixel positions $\mathbf{p}$ and $\mathbf{p}'$.
We denote the displacement-producing module by $g_p$ and formulate it as follows:
\begin{equation}
    \mathbf{p}' = g_p(\theta, \phi)\mathbf{p}.
\end{equation}

We require $g_p$ to express complex transformations such as compositions of rotations and scalings.
Therefore, we implement $g_p$ as a two-layer NN whose weights and biases are generated from $(\theta, \phi)$.
More precisely, we define an NN model, weight generator $\xi(\theta, \phi)$, as follows:
\begin{equation}
    \xi(\theta, \phi) = \{A_1, \mathbf{b}_1, A_2, \mathbf{b}_2\},
\end{equation}
where $A_1 \in \mathbb{R}^{d_m \times 2}$, $\mathbf{b}_1 \in \mathbb{R}^{d_m}$, $A_2 \in \mathbb{R}^{2 \times d_m}$, and $\mathbf{b}_2 \in \mathbb{R}^{2}$.
Then, we compute the displacement $\Delta \mathbf{p}$ as
\begin{equation}
    \Delta \mathbf{p} = A_2 \, r(A_1 \mathbf{p} + \mathbf{b}_1) + \mathbf{b}_2,
\end{equation}
\label{eq:displacement_two_layer_nn}
where $r$ is a rectified linear unit (ReLU) function.
Finally, we acquire the output position as $\mathbf{p}' = \mathbf{p} + \Delta \mathbf{p}$.

\subsection{Constraints on the transformation $g$}

First, we optimize the NN models using a reconstruction loss $\mathcal{L}_r$ so that the transformation $g$ accurately reflects the changes between images $\mathbf{x}$ and $\mathbf{y}$ as follows:
\begin{equation}
    \mathcal{L}_r = MSE(\mathbf{y}, g(\theta, \phi)\mathbf{x}),
    \label{eq:reconstruction_loss}
\end{equation}
where MSE is the mean squared error.

Next, we design loss functions so that the transformation $g$ satisfies the properties of a group.
Regarding the existence of the inverse element, we consider the parameters of the inverse element $(\theta^{-1}, \phi^{-1})$.
They should satisfy $g(\theta^{-1}, \phi^{-1}) \circ g(\theta, \phi) = g(\theta, \phi) \circ g(\theta^{-1}, \phi^{-1}) = e$.
In the case of $g_p$, when it is an identity transformation, the output position $\mathbf{p}'$ is the same as the input position $\mathbf{p}$.
We enforce this property in the space of pixel positions as follows:
\begin{align}
    \mathcal{L}_i = &\ MSE\big(\mathbf{p},\ g_p(\theta^{-1}, \phi^{-1}) \circ g_p(\theta, \phi)\mathbf{p}\big) \notag \\
    &+ MSE\big(g_p(\theta^{-1}, \phi^{-1}) \circ g_p(\theta, \phi)\mathbf{p},\ \mathbf{p}\big).
    \label{eq:inverse_loss}
\end{align}

For two loss functions $\mathcal{L}_i$ and $\mathcal{L}_r$, we obtain the parameters $(\theta, \phi, \theta^{-1}, \phi^{-1})$ from a pair of images $(\mathbf{x}, \mathbf{y})$ using an image-based parameter estimator $E_i$ as follows:
\begin{equation}
    E_i(\mathbf{x}, \mathbf{y}) = (\theta, \phi, \theta^{-1}, \phi^{-1}).
\end{equation}

Regarding the existence of the identity element, we consider the parameters of the identity element $(\theta_e, \phi_e)$.
They should satisfy $g(\theta_e, \phi_e) = e$.
We prepare $(\theta_e, \phi_e)$ as learnable parameters and optimize them using the loss function $\mathcal{L}_e$ as follows:
\begin{equation}
    \mathcal{L}_e = MSE(\mathbf{p}, g_p(\theta_e, \phi_e)\mathbf{p}).
\end{equation}

Regarding the closure property, for two transformations $g(\theta_1, \phi_1)$ and $g(\theta_2, \phi_2)$, there should exist a set of parameters $(\theta_3, \phi_3)$ such that
\begin{equation}
    g(\theta_1, \phi_1) \circ g(\theta_2, \phi_2) = g(\theta_{1,2}, \phi_{1,2}).
\end{equation}
To model this, we provide a position-based parameter estimator $E_p$:
\begin{equation}
    E_p(\{(\mathbf{p}, \mathbf{p}')\}) = (\hat{\theta}, \hat{\phi}),
\end{equation}
where $\{(\mathbf{p}, \mathbf{p}')\}$ is a set of pairs of pre- and post-transformation coordinates.
Because a single $(\mathbf{p}, \mathbf{p}')$ pair does not provide sufficient information to determine the parameters $(\hat{\theta}, \hat{\phi})$, we prepare multiple pairs.
We define the closure loss $\mathcal{L}_{c2}$ as
\begin{equation}
    \mathcal{L}_{c2} = MSE(g_p(\theta_1, \phi_1) \circ g_p(\theta_2, \phi_2)\mathbf{p}, g_p(\hat{\theta}_{1,2}, \hat{\phi}_{1,2})\mathbf{p}),
    \label{eq:oc_loss}
\end{equation}
where $\hat{\theta}_{1,2}$ and $\hat{\phi}_{1,2}$ are the parameters estimated from $g_p(\theta_1, \phi_1) \circ g_p(\theta_2, \phi_2)\mathbf{p}$.
To enhance the effect of this constraint with a finite number of dataset samples, we also consider compositions of more than two transformations, denoted by $g(\theta_1, \phi_1) \circ ... \circ g(\theta_n, \phi_n) = g(\theta_{1,\dots,n}, \phi_{1,\dots,n})$.
We define the closure loss of $n$ transformations $\mathcal{L}_{cn}$ as
\begin{equation}
    \mathcal{L}_{cn} = MSE(g_p(\theta_1, \phi_1) \circ ... \circ g_p(\theta_n, \phi_n)\mathbf{p}, g_p(\hat{\theta}_{1,\dots,n}, \hat{\phi}_{1,\dots,n})\mathbf{p}),
    \label{eq:cn_loss}
\end{equation}
where $\hat{\theta}_{1,\dots,n}$ and $\hat{\phi}_{1,\dots,n}$ are the parameters estimated from $g_p(\theta_1, \phi_1) \circ ... \circ g_p(\theta_n, \phi_n)\mathbf{p}$.

In the proposed method, the transformation $g$ satisfies the associative law, as it is formulated in Eq.~\ref{eq:displacement_two_layer_nn}.
Therefore, we do not need to impose any loss function for the associative law.

\subsection{Homomorphism constraint}
To ensure that the transformation parameter $\theta$ satisfies the homomorphism property in Eq.~\ref{eq:homomorphism_constraint}, we implement the binary operation $\cdot$ on $\theta$ using an NN model.
This NN model takes two parameters $\theta_1$ and $\theta_2$ and outputs the composition result $\theta_1 \cdot \theta_2$.
We use the estimated parameters $\hat{\theta}_{1,2}$ obtained when calculating the closure loss $\mathcal{L}_{c2}$.
Then the homomorphism loss $\mathcal{L}_{h2}$ is defined as
\begin{equation}
    \mathcal{L}_{h2} = MSE(\theta_1 \cdot \theta_2, \hat{\theta}_{1,2}).
    \label{eq:h2_loss}
\end{equation}
Similarly to the closure loss $\mathcal{L}_{cn}$, we consider compositions of more than two transformations.
The homomorphism loss of $n$ transformations $\mathcal{L}_{hn}$ is defined as
\begin{equation}
    \mathcal{L}_{hn} = MSE(\theta_1 \cdot ... \cdot \theta_n, \hat{\theta}_{1,\dots,n}).
    \label{eq:hn_loss}
\end{equation}

\subsection{Constraints on the parameters $(\theta, \phi)$}

Our goal is to obtain a meaningful normal subgroup $N$ from the group $G$.
However, if we impose only the homomorphism constraint, two trivial cases can exist.
One case is $\operatorname{Ker}(f) = G$, where $\theta$ does not change over variations of $g$ and only $\phi$ is used.
The other case is $\operatorname{Ker}(f) = \{e\}$, where $\phi$ does not change and only $\theta$ is used.
Both cases are undesirable, because the resulting categorization of $G$ by $\operatorname{Ker}(f)$ becomes meaningless.
To avoid these trivial cases, both $\theta$ and $\phi$ should vary over variations of $g$.
Therefore, we constrain the variance of each axis of $\theta$ and $\phi$ within a mini-batch to be non-zero.
We encourage the variance to be close to $1$ and define the variance loss $\mathcal{L}_v$ as follows:
\begin{equation}
    \mathcal{L}_v = MSE(1, \mathrm{Var}(\theta)) + MSE(1, \mathrm{Var}(\phi)).
    \label{eq:variance_loss}
\end{equation}

We also consider the uniqueness between the parameters $(\theta, \phi)$ and the transformation $g$.
Because the displacement-producing module $g_p$ is nonlinear, different parameter pairs $(\theta, \phi)$ can potentially result in the same transformation.
Additionally, the parameters estimated from images may contain redundant information beyond what is necessary to specify $g$.
Then, the homomorphism constraint might be satisfied using this redundancy instead of the transformation $g$.
To prevent this, we introduce a uniqueness constraint that enforces the recoverability of the parameters from the corresponding transformation.
The uniqueness loss $\mathcal{L}_u$ is defined as
\begin{equation}
    \mathcal{L}_u = MSE(\theta, \hat{\theta}) + MSE(\phi, \hat{\phi}),
    \label{eq:uniqueness_loss}
\end{equation}
where $\hat{\theta}$ and $\hat{\phi}$ are the parameters estimated from $g(\theta, \phi)$ using the position-based parameter estimator $E_p$.

\subsection{Total loss}

Combining the above losses, the total loss $\mathcal{L}$ is defined as
\begin{equation}
    \mathcal{L}
    = \mathcal{L}_r
    + \alpha \mathcal{L}_i
    + \beta \mathcal{L}_e
    + \gamma \mathcal{L}_{cn}
    + \delta \mathcal{L}_{hn}
    + \epsilon \mathcal{L}_v
    + \zeta \mathcal{L}_u,
    \label{eq:loss_all}
\end{equation}
where $\alpha, \beta, \gamma, \delta, \epsilon, \zeta \in \mathbb{R}$ are weighting coefficients.

In our previous study \cite{Nishitsunoi2025}, we additionally imposed two auxiliary constraints, even though group decomposition theory itself does not require it: (1) assuming an image sequence generated by uniform linear motion and (2) restricting the factor transformations to be isometric.
By contrast, the proposed method directly focuses on general transformations between two images $(\mathbf{x}, \mathbf{y})$ without requiring a sequence or uniform linear motion, and without imposing \emph{a priori} constraints such as isometry.
As a result, the proposed method can be applied to a wider range of scenarios.

\section{Experiments}\label{sec:experiments}

\subsection{Dataset}

The dataset used in this study is based on the \textbf{Syn-obj} dataset, which contains single object images undergoing geometric transformations and was used in our previous study \cite{Komatsu2025}.
In this section, we describe the differences from the original dataset.

We consider a dataset of pairs of images.
The image size is $64 \times 64$ pixels and the object occupies a $32 \times 32$ region.
We transform the object using a combination of two elementary transforms.
First, we apply an \emph{object-centered} transformation and then a \emph{global} transformation.
Object-centered transformation means that the center of the transformation, such as the rotation center or scaling center, coincides with the center of the object.
Global transformation means that the center of the transformation might not coincide with the center of the object.
We prepare three variants of the dataset:
(i) rotation and translation:
(ii) scale and rotation: and
(iii) scale and translation.
In all variants, we draw the magnitudes of the transformations as follows:
rotation angle in $[-40^\circ, 40^\circ]$;
translation in $[0, 16]$ pixels with a random direction; and
scale factor in $[0.7, 1.4]$.
We generate $1,000$ image pairs per dataset variant.

\subsection{NN models and training settings}

We focus on the case in which the number of parameters is two: $g(\theta, \phi)$.
We set both dimensions $d_\theta$ and $d_\phi$ to $4$.

We use the following shorthand to describe the NN models:
$L(i,o)$ is a linear layer with input dimension $i$ and output dimension $o$;
$C_{k,s,p}(i,o)$ is a $k \times k$ convolution with stride $s$, padding $p$, input channels $i$, and output channels $o$;
and $r$ denotes an ReLU activation function.

The weight-generator $\xi(\theta,\phi)$ consists of [$L(8,128)$, $r$, $L(128,8)$].
For stable optimization in the early stage of training, we initialize the weights in this module with small random values sampled from $\mathcal{N}(0, 0.01^2)$ and set the biases to zero.
Regarding the output weights and biases $A_1$, $\mathbf{b}_1$, $A_2$, $\mathbf{b}_2$, we set the middle dimension $d_m$ to $128$.

As the input of the image-based parameter estimator $E_i$, we concatenate an image pair $(\mathbf{x}, \mathbf{y})$ and pixel positions in the channel direction.
We provide two NN models for $\theta$ and $\phi$.
The designs of these two NN models are identical, but the weights are not shared.
First, we process the input using convolutional NNs (CNNs), which consist of [$C_{4,2,1}(8,32)$, $r$, $C_{4,2,1}(32,64)$, $r$, $C_{4,2,1}(64,128)$, $r$].
Then, we conduct average pooling to reduce the spatial map to $4 \times 4$.
Finally, we flatten the pooled feature map and apply it to two linear layers which consist of [$L(2048,256)$, $r$, $L(256,8)$].
We divide the output of the linear layers into components corresponding to the original parameter (e.g., $\theta$) and its inverse (e.g., $\theta^{-1}$).

Regarding the training process, which involves the position-based parameter estimator $E_p$, we sample $64$ pixel positions $\mathbf{p}$.
To capture the feature of the transformation $g$, we sample the positions $\mathbf{p}$ with a grid of size $8 \times 8$.
Along the $x$-axis (and similarly for $y$), we place $8$ points so that the step size is fixed to $8$ pixels.
Additionally, we shift the position of the grid by a random offset in $[0, 8)$ pixels.
As a result, the shape of the sampled positions is $8 \times 8 \times 2$.

The position-based parameter estimator $E_p$ consists of two separate CNNs without shared weights, similar to the image-based parameter estimator $E_i$.
As the input of the position-based parameter estimator $E_p$, we concatenate the sampled input positions $\mathbf{p}$ and output positions $\mathbf{p}'$ in the channel direction.
We process the input using CNNs, which consist of [$C_{4,2,1}(4,32)$, $r$, $C_{4,2,1}(32,64)$, $r$, $C_{2,2,0}(64,4)$, $r$].
We acquire the estimated parameters by flattening the output feature map.

The learnable binary operation of $\theta$ consists of [$L(8,128)$, $r$, $L(128,8)$].

We set the batch size to $100$.
For the closure loss $\mathcal{L}_{cn}$ and homomorphism loss $\mathcal{L}_{hn}$, we randomly sample transformations $g(\theta, \phi)$ from the current batch and prepare $1,000$ compositions per step.
We draw the number of factors in each composition uniformly between $2$ and $8$.
We repeat training with random seeds from $1$ to $40$.
We set the learning rate to $0.001$.
In Eq.~\ref{eq:loss_all}, $\mathcal{L}_r$ has weight $1$, and $(\alpha,\beta,\gamma,\delta,\epsilon,\zeta) = (1, 1, 1, 0.1, 0.1, 0.01)$ for $\mathcal{L}_i$, $\mathcal{L}_e$, $\mathcal{L}_{cn}$, $\mathcal{L}_{hn}$, $\mathcal{L}_v$, and $\mathcal{L}_u$, respectively.

\subsection{Evaluation method}

We determine whether the obtained normal subgroups are appropriate based on evaluation metrics.
We assume the normal subgroup for each dataset as follows:
(i) for rotation+translation datasets, translation only;
(ii) for scale+rotation datasets, scale only; and
(iii) for scale+translation datasets, translation only.
For each image pair in the dataset, we prepare two types of ground-truth transformations: one is the composition of the two factors, denoted by $g_c$, and the other uses only the single factor corresponding to the assumed normal subgroup, denoted by $g_n$.
Then, we test whether $g_n$ acts as the kernel of the homomorphism $f$.

First, we sample the input positions, and then calculate two output positions using $g_c$ and $g_n$.
Next, we acquire two parameters $\theta_c$ and $\theta_n$ using the image-based parameter estimator $E_i$.
If $g_n$ is the kernel of homomorphism $f$, $\theta_n$ should coincide with the identity element $e_\theta$.
Then, we evaluate whether the error $MSE(\theta_n, e_\theta)$ is significantly small.

Finally, we use the normalized score $I_h$ to evaluate the performance of the homomorphism constraint as follows:
\begin{equation}
    I_h = \frac{MSE(\theta_n, e_\theta)}{MSE(\theta_c, e_\theta)}.
\end{equation}
We normalize the error using $MSE(\theta_c, e_\theta)$ because it may also change with or without the homomorphism constraint.

\subsection{Experimental results}

\begin{figure}[t]\centering
\begin{subfigure}[t]{0.32\columnwidth}\centering\fbox{\includegraphics[trim=5 7 5 5, clip, width=0.93\textwidth]{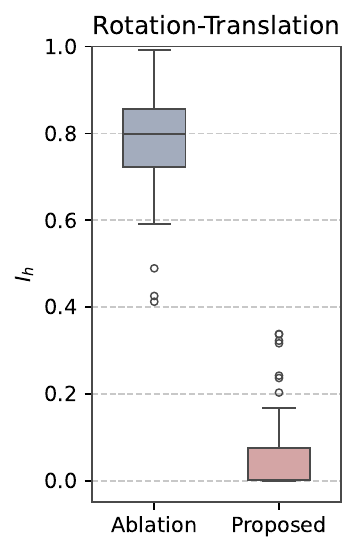}}\caption{}\label{fig:2_a}\end{subfigure}
\begin{subfigure}[t]{0.32\columnwidth}\centering\fbox{\includegraphics[trim=5 7 5 5, clip, width=0.93\textwidth]{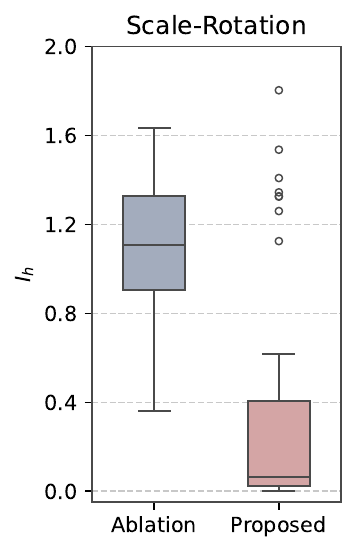}}\caption{}\label{fig:2_b}\end{subfigure}
\begin{subfigure}[t]{0.32\columnwidth}\centering\fbox{\includegraphics[trim=5 7 5 5, clip, width=0.93\textwidth]{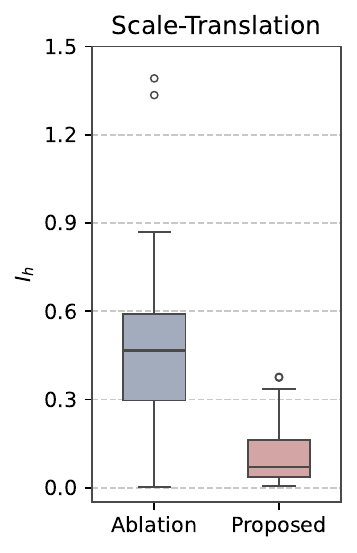}}\caption{}\label{fig:2_c}\end{subfigure}\caption{
Boxplots comparing the evaluation metric $I_h$ with and without the homomorphism loss $\mathcal{L}_h$.
Boxplot \textbf{(a)} is for the dataset variant (i) rotation and translation, \textbf{(b)} is for the dataset variant (ii) scale and rotation, and \textbf{(c)} is for the dataset variant (iii) scale and translation.
}\label{fig:2}\end{figure}

\begin{figure}[t]\centering
\begin{subfigure}[t]{0.96\columnwidth}\centering\fbox{\includegraphics[trim=0 0 85 25, clip, width=1.0\textwidth]{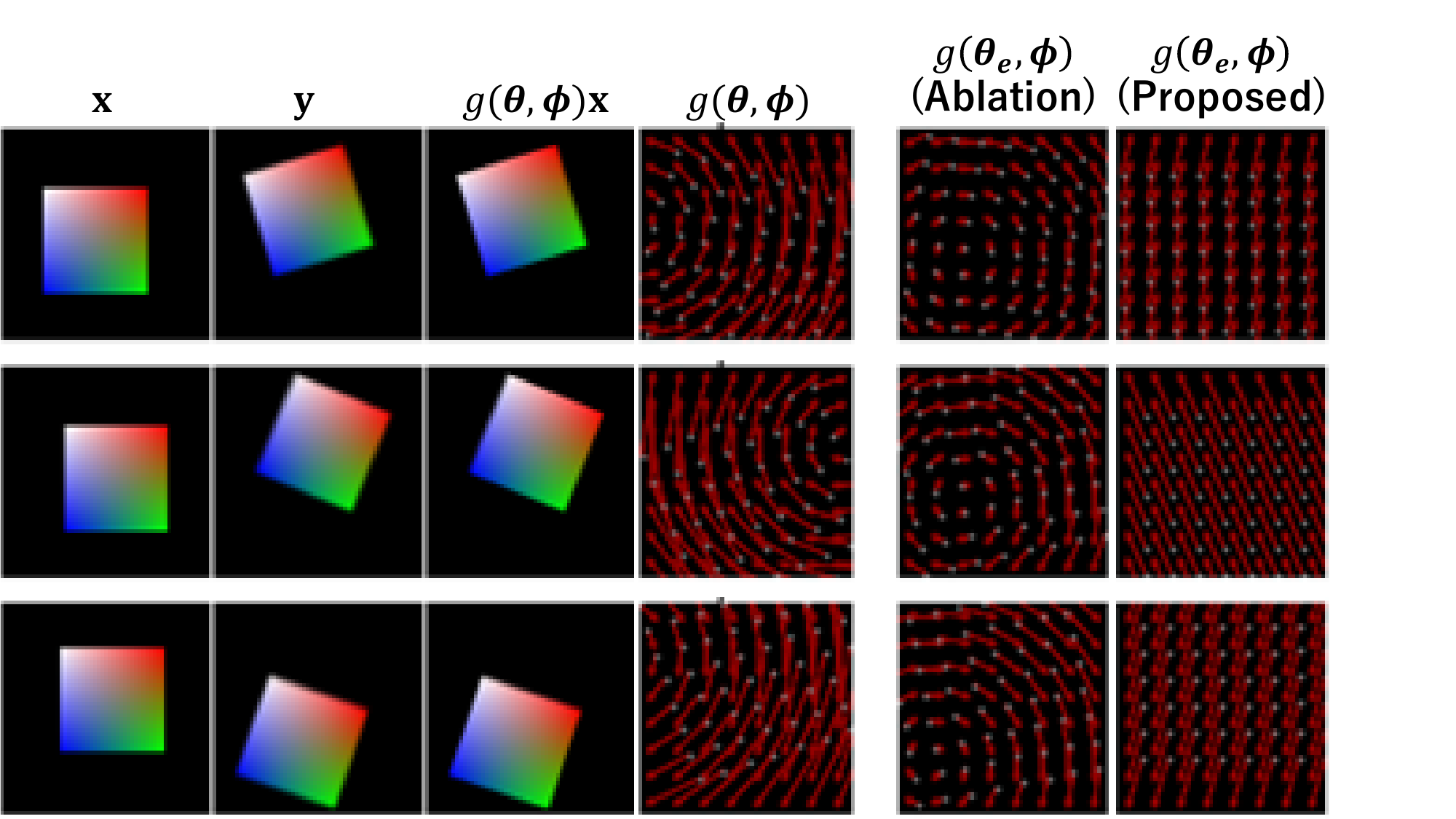}}\caption{Rotation and translation}\label{fig:3_a}\end{subfigure}\vspace{10pt}
\begin{subfigure}[t]{0.96\columnwidth}\centering\fbox{\includegraphics[trim=0 0 85 25, clip, width=1.0\textwidth]{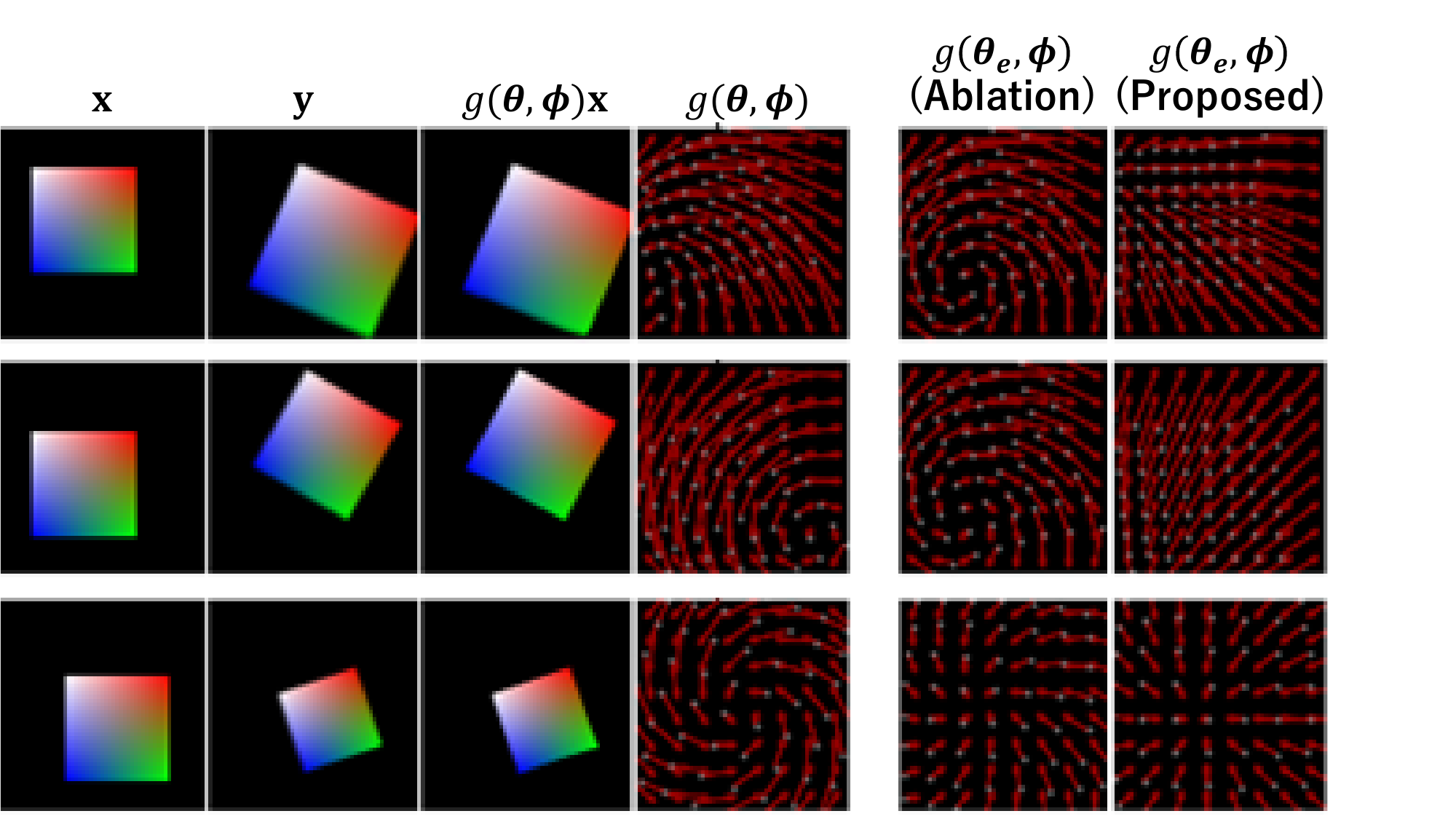}}\caption{Scale and rotation}\label{fig:3_b}\end{subfigure}\vspace{10pt}
\begin{subfigure}[t]{0.96\columnwidth}\centering\fbox{\includegraphics[trim=0 0 85 25, clip, width=1.0\textwidth]{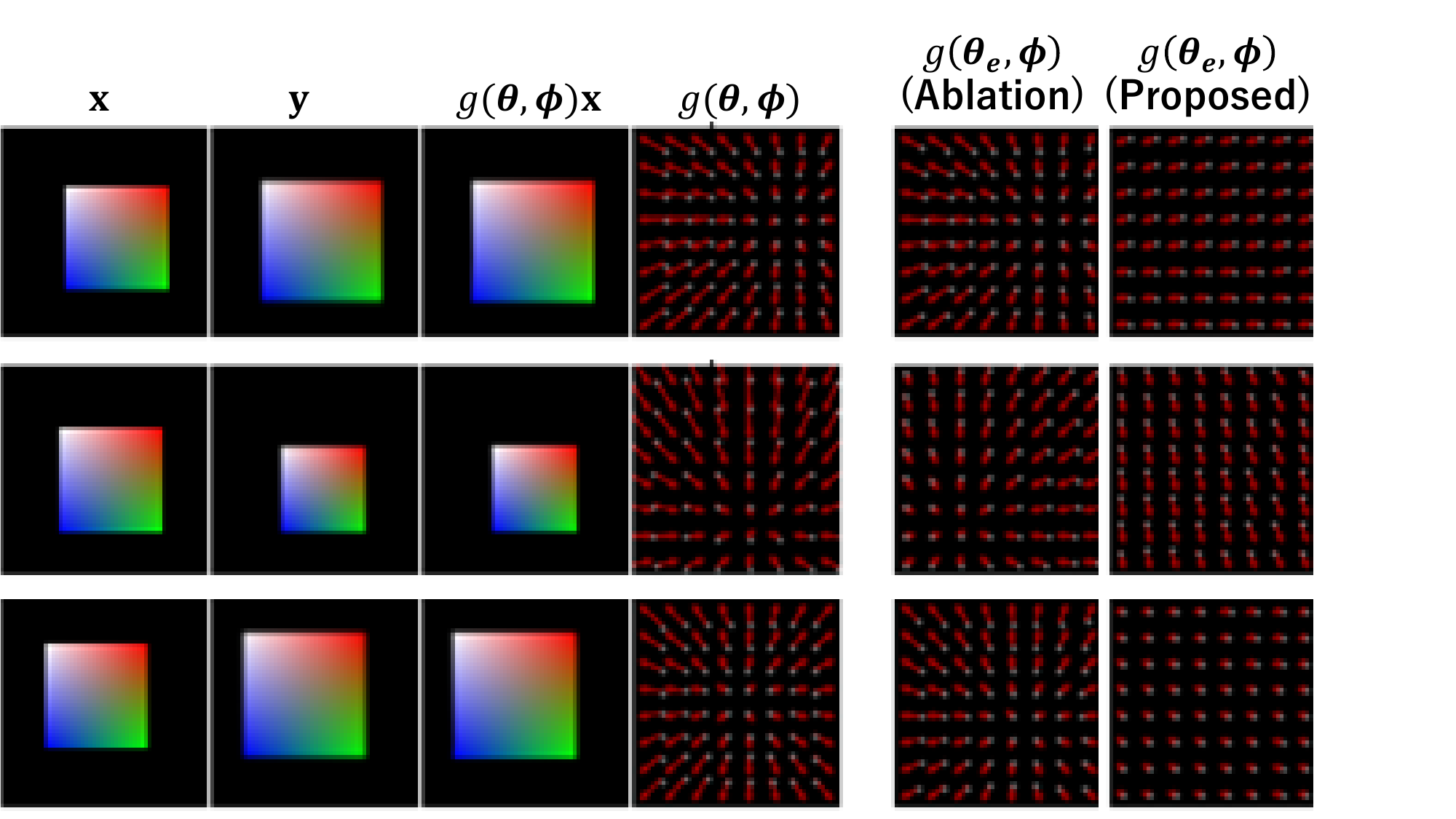}}\caption{Scale and translation}\label{fig:3_c}\end{subfigure}\caption{
Examples of the learning results.
}\label{fig:3}\end{figure}

As an ablation condition, we set the weight $\delta$ of the homomorphism loss $\mathcal{L}_h$ to $0$.
Fig.~\ref{fig:2} compares the evaluation metric $I_h$ with and without the homomorphism loss $\mathcal{L}_h$.
Across all dataset variants, training with $\mathcal{L}_h$ yielded significantly smaller $I_h$ than ablation without it ($p < 0.05$, $t$-test).
This indicates that the proposed homomorphism constraint contributed to ensuring that transformations considered as normal subgroups, $g_n$, were appropriately categorized as the kernel.

Fig.~\ref{fig:3} shows examples of the learning results.
For each dataset variant, the rows in each block correspond to three sampled examples, and the columns (from left to right) show the input image $\mathbf{x}$; target image $\mathbf{y}$; transformed image $g(\theta,\phi)\mathbf{x}$ after learning; transformation $g(\theta,\phi)$; kernel component under ablation, $g(\theta_e, \phi)$; and kernel component with the proposed method, $g(\theta_e, \phi)$.
We obtain the kernel component by combining the identity element $\theta_e$ and the parameter $\phi$ estimated from the input image pair $(\mathbf{x}, \mathbf{y})$.
The red arrows of the transformation $g(\theta,\phi)$ and kernel component $g(\theta_e, \phi)$ indicate the displacement vectors $\Delta \mathbf{p}$ on each pixel position.

The target image $\mathbf{y}$ is consistent with the transformed image $g(\theta,\phi)\mathbf{x}$.
This indicates that the model learned the overall transformation $g(\theta,\phi)$ appropriately.
Under the ablation, the kernel component $g(\theta_e, \phi)$ contained mixed factors from the two transformation components.
By contrast, With the homomorphism constraint, the kernel component $g(\theta_e, \phi)$ contained only the factor corresponding to the transformation assumed to form the normal subgroup $N$.
This indicates that the homomorphism constraint contributed to acquiring the normal subgroup $N$ appropriately.

\section{Conclusion}\label{sec:conclusion}

In this study, we proposed a novel formulation and learning model for transformation categorization based on group decomposition theory.
We revisited homomorphism formulations in prior work that entailed theoretically unnecessary assumptions, and reformulated homomorphism within a framework that decomposes the parameters of a single transformation into multiple factors.
This yielded a formulation free of those unnecessary theoretical conditions.
Furthermore, we successfully proposed a learning model applicable to a broader range of scenarios than earlier approaches, because it does not rely on the a priori constraints they imposed.
We conducted ablation studies on multiple datasets involving translation, rotation, and scale.
The results demonstrated that the homomorphism constraint contributed to acquiring appropriate normal subgroups.
To the best of our knowledge, this is the first study in which group-decomposition-based methods have been quantitatively shown to be effective for appropriately categorizing transformations in scenarios that include non-isometric transformations.

Throughout this study, we targeted the two-level relationship between the transformation group $G$ and a normal subgroup $N \trianglelefteq G$.
However, the proposed framework can plausibly be extended toward learning richer hierarchical structures.
For example, if parameters already governed by a homomorphism constraint are further decomposed and an additional homomorphism constraint is imposed on another part of the factorization, hierarchies with three or more levels may be obtained.
This direction is promising, in part because it may also contribute to computational accounts of how infants acquire rich, structured knowledge.

\bibliographystyle{ieeetr}
{\small
\bibliography{bib/references}}

\end{document}